\documentclass[conference]{IEEEtran}
\usepackage{times}

\usepackage[numbers]{natbib}
\usepackage{multicol}
\usepackage[bookmarks=true]{hyperref}

\usepackage{graphicx}
\usepackage{amsmath}	
\usepackage{amsthm}
\usepackage{amsfonts}
\usepackage{multirow}
\usepackage{algorithm}
\usepackage{algpseudocode}
\usepackage{algorithmicx}
\usepackage{bbold}
\usepackage{caption}
\usepackage{mathtools}
\usepackage{booktabs}
\usepackage{hyperref}
\usepackage[dvipsnames]{xcolor}
\usepackage{float}

\newcommand{\specialcell}[2][c]{%
\begin{tabular}[#1]{@{}c@{}}#2\end{tabular}}
\usepackage{xspace}

\newcommand{\aname}{{\small PROMPT}\xspace}
\newcommand\methodName{PROMPT\xspace}
\newcommand\methodNameFull{Ab Initio Particle-based Object Manipulation}

\renewcommand{\eqref}[1]{(\ref{#1})}
\newcommand{\figref}[1]{Fig.~\ref{#1}}

\newcommand{\tabref}[1]{Table~\ref{#1}}

\newcommand{\eg}{\textrm{e.g.}}
\newcommand{\etc}{\textrm{etc.}}

\newtheorem{definition}{Definition}

\begin{document}

\title{\textrm{Ab Initio} Particle-based Object Manipulation}

\author{
\authorblockN{Siwei Chen,  Xiao Ma, Yunfan Lu and David Hsu}
\authorblockA{National University of Singapore}
\{siwei-15, xiao-ma, luyf, dyhsu\}@comp.nus.edu.sg
}

\twocolumn[{%
\renewcommand\twocolumn[1][]{#1}%
\maketitle
\begin{center}
	\centering
	\begin{tabular}{c@{\hspace*{15pt}}c@{\hspace*{3pt}}c@{\hspace*{3pt}}c}
    \includegraphics[width=0.46\linewidth]{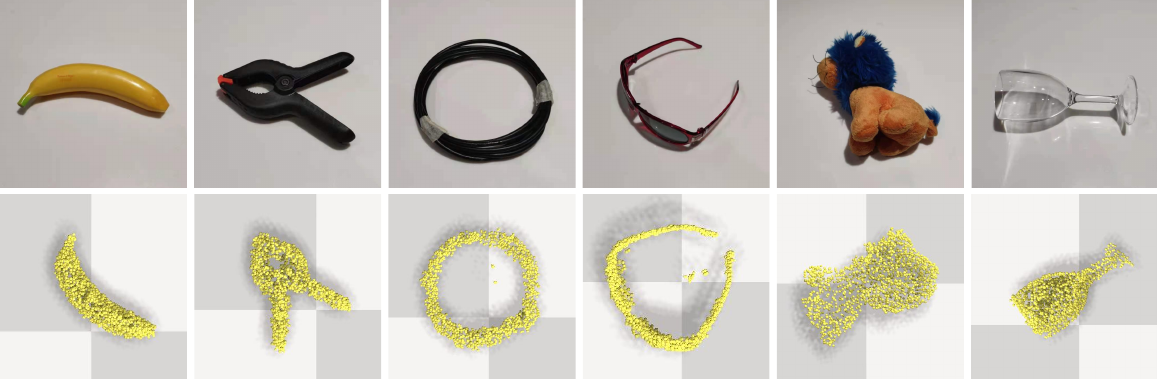} &
	\includegraphics[width=0.15\linewidth]{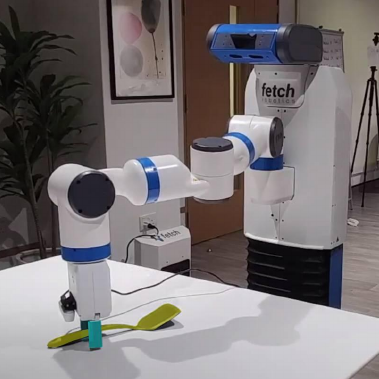} &
	\includegraphics[width=0.15\linewidth]{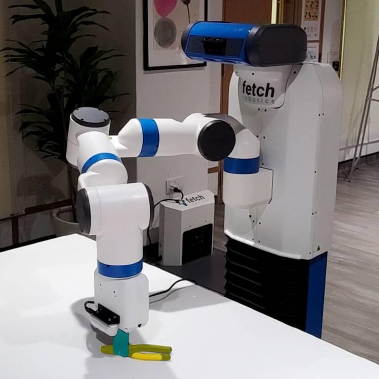} &
    \includegraphics[width=0.15\linewidth]{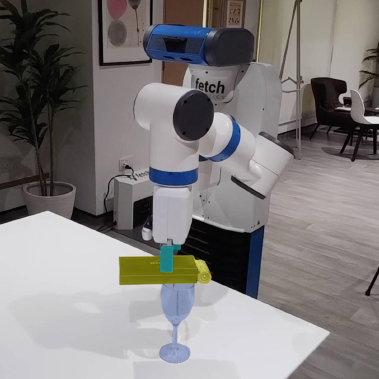}\\
    \footnotesize Reconstructed particle representation of some household objects
    &\footnotesize Grasping
    &\footnotesize Pushing
    &\footnotesize Placing

	\end{tabular}
	\centering
    \captionof{figure}{\aname constructs a particle  representation for object manipulation. 
    }\label{fig:showcase}
\end{center}%
}]

\begin{abstract}
\textit{Particle-based Object Manipulation} (\aname) is a new method for robot manipulation of novel objects, without prior object models or pre-training on a large object data set. The key element of \aname is a  particle-based object representation, in which each particle represents a point in an object, the local geometric, physical, and other features of the point, and also its relation with other particles. The particle representation  connects visual perception with  robot  control. Like  data-driven methods, \aname 
infers the object representation online in real time from the visual sensor.
Like  model-based methods, \aname leverages the particle representation to reason about the object's geometry and dynamics, and  choose suitable manipulation actions accordingly.   \aname thus combines the strengths of model-based  and data-driven methods. We show empirically that \aname successfully handles a variety of everyday objects, some of which are transparent. It handles various manipulation tasks, including grasping, pushing, \etc\,. Our experiments also show that \aname outperforms a state-of-the-art data-driven grasping method on everyday household objects, even though it does not use any offline training data. The code and a demonstration video are available online\footnote{The code and a demonstration video are available online at
 \url{ https://adacomp.comp.nus.edu.sg/2021/06/26/particle-based-robot-manipulation}.}

\end{abstract}

\IEEEpeerreviewmaketitle

\section{Introduction}

Object manipulation is a central question of   robotics research. It embodies a broad spectrum of challenges from robot design to sensing and control algorithms. One key challenge of robot manipulation systems is object representation, which connects robot sensing and control. Take, for example, grasping.  Classic model-based methods capture the object's geometric and physical properties in a compact analytic model, which enables sophisticated {reasoning} about the grasp quality, through quasi-static or dynamic analysis. However, acquiring such an object model from sensor data is often difficult. Also, simplifying assumptions and inaccuracies in the models often result in brittle grasps that fail during the robot execution. In contrast,  data-driven methods learn the grasps directly from prior experiences. They focus on the perception  of object features critical for predicting good grasps and are often more robust in practice. The data-driven methods, however, find difficulty in generalizing to novel objects, unseen in the training data.
This work aims for a general, flexible object representation that enables both effective construction  from sensor data  and sophisticated reasoning of geometry and dynamics for manipulation, thus achieving the benefits of both  data-driven and model-based methods.  


Specifically, we introduce \textit{Particle-based Object Manipulation} (\aname).
The key element of \aname is a particle-based object representation, in which each particle represents a point in the object, the local geometric, physical, and other features of the point, and also its relation with other particles. The particle representation is general and approximates  objects with diverse geometric shape and visual appearance (\figref{fig:showcase}).  Given a set of  RGB images of an object from multiple camera views, \aname first constructs a particle representation of the object  \textit{online} in real time.
For each camera view, it projects the particles into the image plane and matches with the foreground-segmented object in the image, using an improved Chamfer distance measure. 
\aname  then uses the reconstructed particle set as an approximate state representation of the object for reasoning.  It performs particle-based dynamics simulation to predict the effects of manipulation actions on the object and chooses  the desired 
actions through model
predictive control. \aname is capable of a  variety of  manipulations tasks: grasping, pushing, placing, \ldots (\figref{fig:showcase}).

Our experiments show that \aname successfully constructs the particle representation for a wide variety of everyday household objects, including some that are transparent and  particularly  challenging for reconstruction from visual sensor data. \aname outperforms Dex-Net 2.0~\cite{mahler2017dex}, a state-of-the-art data-driven method for object grasping. It compares favorably with previous work~\cite{Li2018PushNet} on object pushing.  Finally, it manipulates an object with unknown mass distribution, grasps it, and balances it on a small stand.

\aname is a simple idea, but works surprisingly well. It benefits significantly from the particle representation, which connects visual perception with dynamics reasoning.
The particle representation is simple and general. It allows us to leverage the dramatic progress in particle-based dynamics simulation. 
Compared with  model-based methods, \aname learns the object  representation  automatically from visual sensor data. Compared with  data-driven methods, it requires no offline training data and ``generalizes'' to novel objects trivially. \aname also generalizes  across many different rigid-body manipulation tasks.

\section{Related Works}


\subsection{Manipulation with 3D Mesh Model}
Classical analytic methods for object manipulation often assume a known 3D mesh model of an object and evaluate some specific metrics for action planning \cite{siciliano2008springer}. For example \citet{pokorny2013classical} extends the popular $L^1$ grasp quality measure that considers the ability to resist wrenches in object grasping tasks. Numerous works follow this approach to retrieve the object's 3D mesh model, estimate the pose and plan for actions \citep{ding2000computing, goldfeder2007grasp,  roa2012power,huebner2009grasping, brook2011collaborative}. 
However, given an unseen novel object with no available mesh model, classical methods often fails. To fix this issue, 3D mesh reconstruction with laser scanners are introduced~\cite{marton2010general, marton2011reconstruction,bone2008automated,bohg2011mind}. They consider an object as a combination of a construction of primitive parts, \eg boxes and cylinders.
Nonetheless, generalizing 3D mesh-based models to arbitrary shaped and translucent objects remains challenging.

\subsection{Manipulation with Latent Representation}
Latent state representation closely connects to modern data-driven approaches.
Sensory observations are encoded into latent states and directly mapped to a robot policy by a parameterized function, which is trained from pre-collected labeled data~\cite{mahler2017dex}.
Pioneer works use human labelled data for training~\cite{balasubramanian2012physical, kappler2015leveraging, herzog2014learning, lenz2015deep}. Self-supervised trial and error on real robots provide an alternative to human data labeling~\cite{montesano2012active, oberlin2018autonomously}. Large-scale real robot experiments \cite{pinto2016supersizing, levine2018learning, kalashnikov2018qt} are conducted to collect 
thousands of
grasping experiences across few months. In contrast to training a policy with real-world data, other works train policy in simulators~\cite{viereck2017learning,tobin2018domain}. Specifically, Dex-Net 2.0~\cite{mahler2017dex} trains a grasp quality CNN network over a 6.7 million synthetics depth images grasping dataset and achieves 80\% success rate without any real-world data. Similarly, PushNet \cite{Li2018PushNet} learns a pushing policy entirely in simulation and succeeds in all real robot pushing tasks.

Despite the success of data-driven approaches, the data-driven approaches still suffer from the following limitations: 1) Data gathering: data gathering in the real world is expensive and time-consuming. 2) Distribution Shift: the testing data may differ significantly from the training data, especially when the policy is trained in a simulator and applied in the real world. 3) Dynamics reasoning: the trained policy often lacks explicit dynamics reasoning and may result in sub-optimal actions.

\subsection{Particle Representation \& Reconstruction}

Structure From Motion (SFM) \cite{koenderink1991affine} tracks the feature correspondences cross multi-view images and reconstruct a scene point cloud.  However, the reconstructed point cloud is sparse and incomplete as the point cloud relies on tracked features that can be very sparse and noisy. Other approaches \cite{li2015detail, savinov2016semantic, semerjian2014new} related to multi-view stereo achieve high accuracy and completeness in reconstruction but require time ranging from minutes to even hours and is hence not practically useful in real robot tasks. Shapes from silhouettes (SFS) performs reconstruction from multi-view silhouettes \cite{laurentini1994visual, laurentini1995far}. However, typical SFS, such as the polyhedral approach, often requires perfect silhouettes and problems arise when silhouettes contain errors \cite{franco2008efficient, szeliski1993rapid}. Moreover, it does not scale well as the number of images increases. In our experiments, we need to process dozens to hundreds of images online, and the masking images often contain errors.

    
    


We propose a new online approach for constructing an particle representation and use it for model-based planning. Instead of reconstructing the 3D mesh model using existing approaches that can be slow, our method can quickly achieve good point cloud reconstructions on various objects. On top of that, particle representation connects the visual perception to the known particle-based dynamics model and enables explicit reasoning for different robot actions. Furthermore, particle representation is general and has better potentials. For example, it can model deformable objects and liquid such as a bowl of porridge and a cup of coffee by using different types of particles.

\section{Particle-based Object Manipulation}

\begin{figure*}[ht]
\centering
\includegraphics[width=0.9\linewidth]{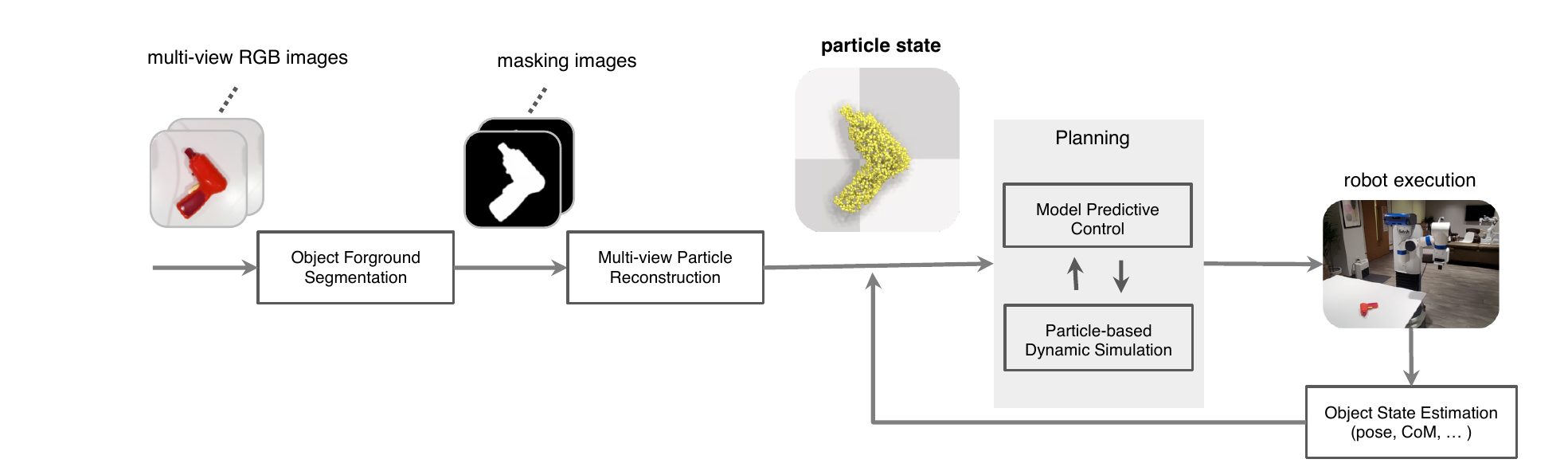}
\caption{Overview. Particle Model-based Grasping takes in multi-view RGB images and reconstruct a 3D particle state which is used for planning with model-predictive control with a state-of-the-art particle-based physics engine. }
\label{Fig:structure}
\end{figure*}

\subsection{Overview}
Our approach aims to generate a universal particle state representation for ab initio rigid object manipulation, such as grasping, pushing, and placing tasks. 
The input to the system is defined as a set of tuples $(M_i, E_i)$, where $M_i$ is the $i_{th}$ masking image and $E_i$ is the $i_{th}$ camera extrinsic matrix. The camera's intrinsic matrix is known. We also assume we can have a sufficient coverage of views on the object and each object is manipulated in isolation.  We define particle state below:

\begin{definition}
\textbf{Particle} We define a particle as $v = (x,y,z, \dot{x}, \dot{y}, \dot{z}, m)$ with position $x,y,z$, velocity $\dot{x}, \dot{y}, \dot{z}$, and mass $m$. A particle represents a point of the object, which distributes on either the surface or the interior of an object.
\end{definition}


\begin{definition}
\textbf{Particle State} We define a particle state as $S = (V,R, \mu)$, where $V$ is the set of object particles defined above and $R$ is a set of direct edges. Each edge $r = (v_1, v_2, a)$ represents the relations $a$ from particle $v_1$ to $v_2$ (e.g. collision, spring connection). $\mu$ refers to the system friction coefficient.
\end{definition}

The particle state definition applies to both rigid and deformable objects, e.g., liquid and cloth, by setting different edge types $R$. In this work, we focus on the rigid objects and leave deformable objects for future study. Specifically, we manually set edge types to specify a rigid object.


Given a particle state, a decision making system outputs an action, which can be adjusted for different tasks. For example, in a top-down object grasping task, we define action space as $A = (x,y,z,\theta )$, with the grasping center as $(x,y,z)$ and the anti-clockwise angle $\theta$ from the positive $x$ axis. Similarly, for an object pushing task, the action space could be defined as $A = ( x_0,y_0,x_1,y_1 )$, which refers to the start and end position of a horizontal push action with a fixed height.


With the particle state representation, we factorize a manipulation problem into two consecutive downstream tasks: 1) accurate estimation of a particle state; 2) effective motion planning with the predicted particle state.


We introduce a new approach called \methodName: \methodNameFull. The high-level structure can be found in \figref{Fig:structure}. The robot first takes RGB images from different view angles, and produces segmentation masks $M$. Given the masking images, our approach predicts the particle state online, focusing on 3D shape reconstruction for the unseen objects. The predicted particle state is then fed into a particle-based simulator, which predicts the future particle states given the current state and an action. With the particle state and the dynamics model, we perform cross-entropy model predictive control (MPC) \cite{richards2005robust} to plan for the best action. The system then executes the best action on the real robot. After each real robot execution, we collect new images and use the observation to update the hidden properties of the object, e.g., mass distribution and position.

\subsection{Multi-view Particle State Estimation}
Data-driven methods learn to reconstruct point cloud by offline training with labeled data. Reconstructing an accurate point cloud for arbitrary objects is extremely challenging due to the limited available training data. However, we argue that recovering an accurate point cloud online for a specific object is much easier. In practice, we observe that with a small amount data, e.g., a few seconds of video, an accurate point cloud can be estimated.

Our approach differs from the traditional structure from motion (SFM). SFM relies on tracked image features that potentially lead to a sparse point cloud, and many essential geometry details are lost. In contrast, our approach generates a dense point cloud to capture all the critical details for particle-based dynamics modeling. We use $N$ particles in the experiment. The original particle $v = (x,y,z, \dot{x}, \dot{y}, \dot{z}, m)$ consists of 7 attributes.  However, at the reconstruction stage, we assume the object's mass is evenly distributed, and the object remains static. We hence simplify a single particle into $\hat{v} = (x,y,z)$. The particle generator function is shown below:

\begin{align}
    G: h \to \hat{V}
\end{align}
where G is the particle generator parameterized by a neural network we aim to train. $h$ is a single number, and $\hat{V}$ are the target particles. A quick idea to train the generator G is to re-project the 3D particles $\hat{V}$ into 2D points on the image plane, and minimize the masking image reconstruction loss. The target particle states can then be achieved by iterative gradient descent with the loss function specified below.

\begin{align}
    \min \sum_{M_i} D(f_{proj}(\hat{V}), M_i)
\end{align}
where $M_i$ refers to the $i_{th}$ masking image, $f_{proj}$ projects the particles from 3D into 2D points (x, y) on the image, and D measure the distance between the re-projected masking image and the ground truth masking image. Simple L1 or L2 loss between the re-projected masking images and ground truth masking images fails. It is because there is no gradient flows through the pixel value of the re-projected masking images, and only the coordinates of the mask are computed from $\hat{V}$.


In fact, measuring the distance $D(f_{proj}(\hat{V}), M_i)$ is equivalent to measure the distance between two non-ordered sets $\{p'\}$ and $\{p\}$, where $p'_n = (x'_n, y'_n) = f_{proj}(\hat{v}_n)$ is the $n_{th}$ re-projected point from $\hat{v}_n$ on the image, and $p_n = (x_n,y_n)$ is the $n_{th}$ point sampled from ground truth masking image $M_i$. Since $\{p\}$ may have a larger size than re-projected set $\{p'\}$, for computational efficiency, we uniformly sampled the same number of the points from $M_i$ to construct $\{p\}$. Figure \ref{fig:masking} shows an example of $\{p\}$ and $\{p'\}$. 

To effectively compute the loss, Chamfer distance is often used to measure the average distance between two sets. Original Chamfer distance $D_{CD}(\{p'\}, \{p\})$ has the form:
\begin{figure}[t]
	\centering
	\begin{tabular}{c c c}
    \includegraphics[width=0.15\linewidth]{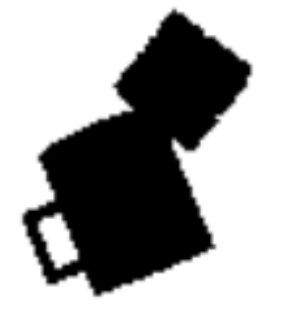} &
	\includegraphics[width=0.15\linewidth]{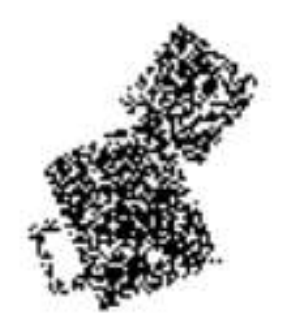} &
    \includegraphics[width=0.15\linewidth]{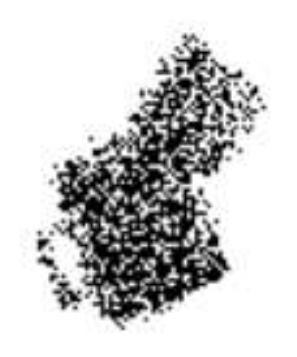}\\
    \footnotesize Binary Masking $M_i$ &\footnotesize Sampled Points $\{p\}$ &\footnotesize Reprojected Points $\{p'\}$

	\end{tabular}
	\centering
	\caption{An example of $\{p\}$ and $\{p'\}$. A cup and a cube lay on the table. Left is one of the ground truth object masking images. The center image shows a set of points $\{p\}$ sampled from left. Right shows a set of points $\{p'\}$ re-projected from particles $\hat{V}$.}
\label{fig:masking}
\end{figure}

\begin{align*}
\sum_{ p' \in \{p'\}} \min_{p \in \{p\}} ||p' - p||^2_2 + \sum_{ p \in \{p\}} \min_{p' \in \{p'\}} ||p' - p||^2_2 
\end{align*}

\begin{figure}[t]
	\centering
	\begin{tabular}{c  c}
    \includegraphics[width=0.3\linewidth]{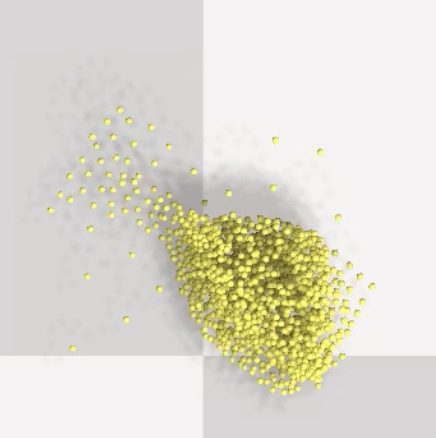} &
    
    \includegraphics[width=0.3\linewidth]{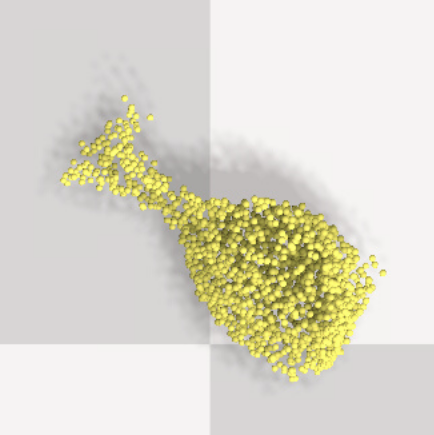}\\
    \small (a) $K=1$ &\small (b) $K =100$
	\end{tabular}
	\centering
	\caption{\small A real-world reconstruction example, a wine glasses, with noisy masking images as input. a) When $K = 1$, it is the standard Chamfer distance, and the reconstruction gets stuck in a local minimum. The algorithm always selects repeating nearest points to optimize. b) $K = 100$, the $K$ nearest Chamfer distance mitigates the problem and provides better results. }
	\label{fig:KNN CD}
\end{figure}

However, we observe that the original Chamfer distance can be stuck in a local minimum. From the definition, each point only finds the closest point from other point sets. This pairing is not unique and can be problematic. For example, in \figref{fig:KNN CD} (a),  standard Chamfer distance falls in a local minimum. If $\{p'\}$ only occupies part of $\{p\}$, the first term $\sum_{ p' \in \{p'\}} \min_{p \in \{p\}} ||p' - p||^2_2$ has a low loss as every $\{p'\}$ finds a point nearby. However, each $p \in \{p\}$  selects the repeating nearest point in $\{p'\}$ to optimize, and all other points are not involved. After few iterations, $\{p'\}$ squeezes in a local area with very few points spread out.

To address this issue, we proposed a new variant of the Chamfer distance called K nearest neighbors Chamfer loss $D_{kN\_CD}(\{p'\}, \{p\})$, defined in \eqref{KNN}. $KN$ selects the K nearest neighbors that belong to the other set. When K is not small, \eg 100, the proposed loss penalizes those points with sparse neighbors and pushes the generated points to have better coverage. \figref{fig:KNN CD} (b) shows an example with $K = 100$ helps to mitigate the local minimum problem.

\begin{align}
\frac{1}{K} \sum_{ p' \in \{p'\}} \sum_{p \in KN} ||p' - p||^2_2 
+ \frac{1}{K} \sum_{ p \in \{p\}} \sum_{p' \in KN} ||p' - p||^2_2 
\label{KNN}
\end{align}

After reconstruction, flying particles are filtered away by a threshold of average neighbour distance due to noisy masking images.  Our approach to reconstructing the point cloud has several advantages: (1) It requires no offline dataset. Hence, we save the efforts from collecting data, which is non-trivial, expensive, and time-consuming. (2) Without using any offline dataset, we prevent over-fitting to existing training data. 
(3) Robust to Translucent Objects. In contrast to the depth sensor that fails on transparent objects, our system works on RGB images to detect the object masking and reconstruct the 3D shape. 

\subsection{Model-based Planning}
Traditional approaches sample a set of action proposals and rank them by handcrafted heuristic metrics. Instead of handcrafting heuristic metrics, we connect the generated particles into the particle-based simulator, building upon a particle-based physics engine. Other existing simulators such as MuJoCo \cite{todorov2012mujoco}, and PyBullet \cite{coumans2019} use the 3D mesh representation, which is difficult to obtain in the test time from the real world. 

To explain the planning algorithms in detail, we use the grasping task as an example. Our approach leverages the particle-based physics engine to perform the virtual grasping and evaluate the final performance based on the two following criteria: (1) Whether the grasping is successful. We define the success criteria as to whether the object is above a height threshold. (2) How robust the grasp is. This metric is defined by the sum of displacement of all the particles along the x-axis and y-axis. The combined reward function is shown below:
\begin{align}
    \textrm{Return} = \mathbb{1} * -10^6 - \sum_{ (x,y,z) \in \hat{V}} ||x - x_0||_1 + ||y - y_0||_1
\end{align}
where $\mathbb{1}$ is the indicator function which suggests whether the grasp is successful or not, $\hat{V}$ is the set of particles predicted, $(x,y,z)$ refers to the 3D position of a particle, and $(x_0,y_0,z_0)$ is the initial position of the same particle.

Given the predicted particle state, the dynamics model, and the reward function, we perform model-based planning that explicitly reasons the dynamics. For simplicity, we use a cross-entropy model-predictive control (MPC)~\cite{richards2005robust} with a uniform prior based on the predicted particles. Instead of randomly sample from the action space, we initialize actions near the predicted particles with random approaching angles and without collision. In each iteration, the system samples sequences of actions from a Gaussian distribution, compute the accumulative reward, update the mean of the Gaussian distribution with the $K$ action sequences with the highest-achieved reward. The fist action of the best action sequence is used as the output.


\algdef{SE}[SUBALG]{Indent}{EndIndent}{}{\algorithmicend\ }%
\algtext*{Indent}
\algtext*{EndIndent}
\begin{algorithm}
    \caption{Grasping Planning with Cross Entropy MPC}
    \begin{algorithmic}[1]
        \Require H \quad  Planning horizon \par
           \quad I \quad Optimization Iteration  \par
           \quad J \quad Candidates per Iteration \par
           \quad K \quad Number of Top candidates to Track \par
           \quad $s = \{o\}$ \quad  Predicted Particle State \par
           
        \Ensure The best estimated action
        
        \State Initialize action $q(a_{t:t+H})$ with a prior distribution
        \For{ optimization iteration $i = 1..I $}
        \Statex \quad  \quad \# Evaluate J actions' performance

        \For{ candidate action $j = 1..J $}
        \State $a^j_{t:t+H} \sim q(a_{t:t+H})$
        \State $s^j_{t:t+H+1} \sim \prod_{\tau = t+1}^{t+H+1}  p(s_\tau | s_{\tau - 1}, a^j_{\tau - 1} )$
        \State $r^j = \sum_{\tau = t+1}^{t+H+1} R(s^j_\tau, a^j_tau)$
        \EndFor
        
        \Statex \quad \quad \# Then select the best K actions to update the action distribution
        \State $ \kappa = \gets argsort(\{r^j\})_{1:K} $
        \State $u_{t:t+H} = \frac{1}{K} \sum_{k \in \kappa} a^k_{t:t+H}$
        \State $\sigma_{t:t+H} = \frac{1}{K-1} \sum_{k \in \kappa} |a^k_{t:t+H} - u_{t:t+H}|$
        \State $q(a_{t:t+H}) \gets Normal( u_{t:t+H}, \sigma_{t:t+H})$
        \EndFor
        \State \textbf{Return} the action mean $u_t$
    \end{algorithmic}
        \label{alg:MPC}
\end{algorithm}

\subsection{Object State Estimation}

For one-shot object grasping, a open-loop planner is able to give a robust policy with careful tuning. However, it is insufficient for manipulation tasks with consecutive actions and unknown physical properties, e.g., object pushing. Specifically, in a pushing task, the robot has to estimate the position and orientation of the object after each push. Moreover, for an object with uneven mass distribution, e.g., a hammer, correctly estimating the center of mass (CoM) is essential for a successful pushing. PROMPT implements close-loop state estimation by dynamically updating the properties of a particle state through the interactions with the object. As illustrative examples, we demonstrate \methodName can be applied to object pose estimation and CoM estimation during a pushing task.


\subsubsection{Closed-loop Object Pose Estimation}  We fix the particle state obtained from the reconstruction stage and only varies the pose parameters $(x, y, \theta)$, which represents $(x,y)$ position on the table and orientation $\theta$. The estimation is done by finding the best pose parameters to minimize the masking image reconstruction loss on new observed image. Similar to CE MPC, we use cross entropy estimation to search for the best pose. With the particle representation, many other optimization techniques can also be used such as particle filter, Bayesian optimization and gradient descent.

\subsubsection{CoM Estimation}
We define the CoM as a Gaussian distribution centered at $(x_c,y_c)$ with variance $\Sigma$. Similar to cross-entropy MPC Algorithm \ref{alg:MPC}, cross-entropy (CE) Estimation is used to find the best parameters $(x_c,y_c)$ and the variance $\Sigma$ that minimizes the gap between the predicted particle state from the particle simulator and the observed state from the real world. In each iteration, the system samples a set of $(x_c,y_c)$ and passes it into the particle-based simulation with all other things fixed, such as the particle states and actions performed. We compare the simulated results with the observed object configuration to select the top candidates. After that, the system recomputes the mean and variance of the CoM Gaussian distribution based on the top candidates and starts a new iteration.


    
    

\begin{table}[]
\centering
\caption{\small 3D Reconstruction Results }
\begin{tabular}{@{}cccccc@{}}
\toprule
            & \multicolumn{2}{c}{Other Baselines} & \multicolumn{3}{c}{PROMPT (Ours)} \\ \cmidrule(l){2-3} \cmidrule(l){4-6} 
            & Point \cite{fan2017point}           & 3D-R2N2 \cite{choy20163d}          & CNN-100    & FC-100    & FC-1     \\ \midrule
Plane       & 0.601           & 0.561             & 0.718      & 0.692     & 0.664    \\
Lamp        & 0.462           & 0.421             & 0.525      & 0.467     & 0.438    \\
Fire arm    & 0.604           & 0.600             & 0.721      & 0.701     & 0.589    \\
Chair       & 0.544           & 0.550             & 0.521      & 0.492     & 0.486    \\
Watercraft  & 0.611           & 0.61              & 0.686      & 0.680     & 0.645    \\ \midrule
Speaker     & 0.737           & 0.717             & 0.509      & 0.540     & 0.587    \\
Couch       & 0.708           & 0.706             & 0.502      & 0.546     & 0.606    \\
table       & 0.606           & 0.580             & 0.517      & 0.490     & 0.493    \\ \midrule
Average IoU & 0.609           & 0.593             & 0.587      & 0.576     & 0.563    \\ \bottomrule
\end{tabular}
\label{table:reconstruction_results}
\end{table}

\begin{figure}[t]
\centering

\includegraphics[width=\linewidth]{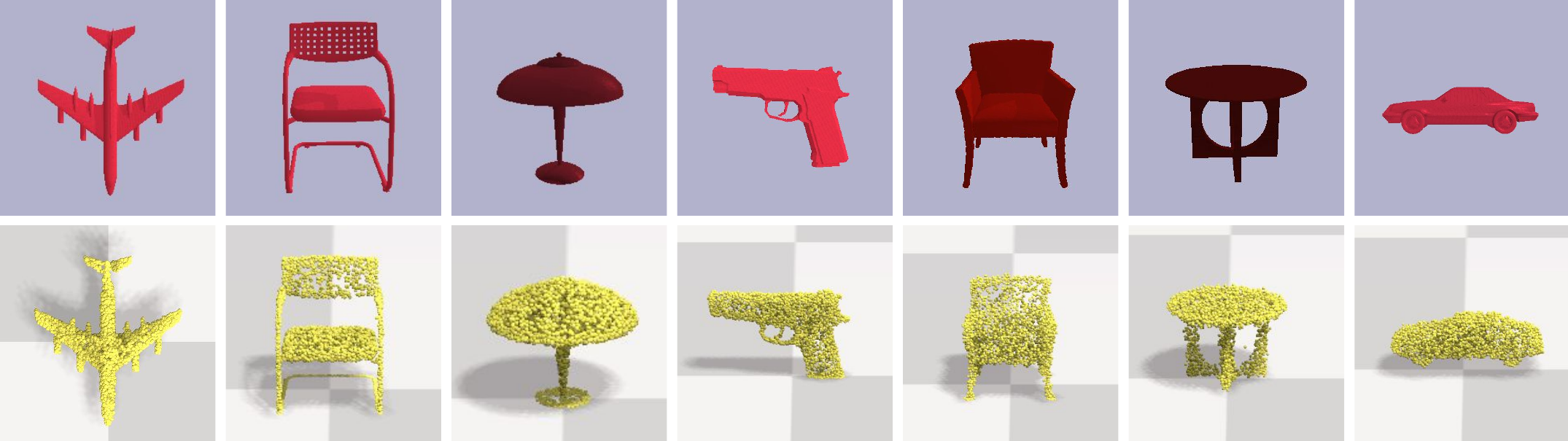}
\label{fig:3d_examples}

\caption{3D Reconstruction Examples. We evaluate our reconstruction performance on the ShapeNet \cite{chang2015shapenet} dataset. There are three variants of our approach. \methodName CNN-100 uses a CNN encoder structured generator with K=100 for K nearest Chamfer distance. FC-100 uses a generator that only consists of a fully-connected layer. FC-1 is an FC layer generator with standard Chamfer distance (K=1). We do not claim to outperform the two baselines but to show we can reconstruct the novel objects reasonably well for next-stage object manipulation. We cite Point Set \cite{fan2017point} and 3D-R2N2 \cite{choy20163d} as a performance indicator. }
\end{figure}

\section{EXPERIMENTS}
We first evaluate our particle reconstruction approach on the dataset ShapeNet \cite{chang2015shapenet} to find out whether our approach can reconstruct the objects well. Next, we compare with Dex-Net 2.0~\cite{mahler2017dex}, a state-of-the-art data-driven method for object grasping without any real grasping data. On top of that, we compare with previous work PushNet \cite{Li2018PushNet} for object pushing.  Finally, the system manipulates an object with unknown mass distribution, pushes it, grasps it, and stacks it on a small stand.

 In summary, our experiments aim to answer the following questions: 1) Can we generate good quality particle reconstructions? 2) How does our approach \aname perform on a real robot grasping task? 3) Can our particle-based framework generalize to other robot tasks such as object pushing and placing tasks?

\begin{figure}[t]
\centering
\includegraphics[width=\linewidth]{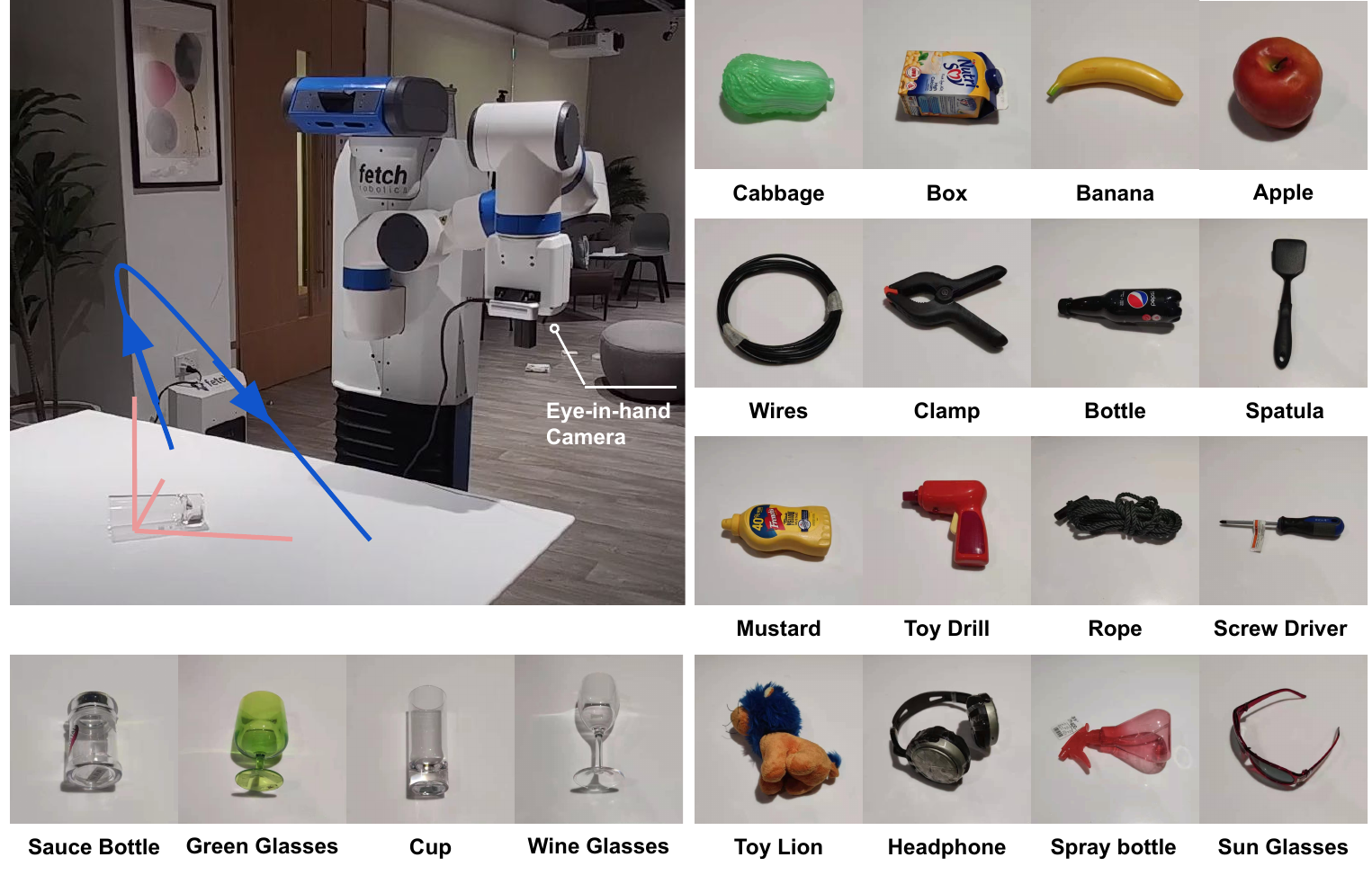}
\caption[HMM]{Top left: a Fetch Robot with an eye-in-hand RGB camera for object grasping. The blue line indicates the trajectory of the eye-in-hand camera during the reconstruction phase. the RGB camera is always pointing at the working area and the view set follows the three-view convention to achieve a good coverage. The top right shows simple items; the center right shows everyday items; the bottom right shows difficult items and the bottom left shows translucent items. Each item is grasped separately.}
\label{fig:robot}
\end{figure}

\subsection{3D Reconstruction}

This section aims to find out whether our 3D reconstruction approach works on complicated, arbitrary-shaped objects. The reconstruction is conducted with 200 views and 4000 particles with a single 2080Ti GPU. We propose three variant implementations of the point cloud generators: FC-100, CNN-100, and FC-1. FC-100 is a single fully connected layer network that takes in a random number and outputs the $N * 3$ sized point cloud with K =100 for the proposed K nearest Chamfer distance. FC-1 is the same FC generator with standard Chamfer distance (K=1). CNN-100 is an auto-encoder structured neural network generator that consists of 12 convolution layers followed by an LSTM layer for a sequence of image inputs and 2 fully connected layers for point cloud output. The variant CNN-100 is used here to determine whether taking in a sequence of image inputs helps with the reconstruction performance.   

\begin{table}[t]
\centering
\caption{\small Robot Grasping Results}
\fontsize{7}{7}\selectfont
\begin{tabular}{clccccc}
\toprule

& & \multicolumn{2}{c}{DexNet 2.0 \cite{mahler2017dex}} & \multicolumn{2}{c}{\methodName (Ours)} \\
\cmidrule(lr){3-4}\cmidrule(lr){5-6}
                 &  & Count    & Suc Rate & Count  & Suc Rate   \\
\midrule

\multirow{4}{*}{\specialcell{Simple \\ Items}} & Banana & 5/5 & \multirow{4}{*}{0.8} & 5/5 & \multirow{4}{*}{0.95} \\

& Box & 2/5 & & 4/5 & \\
& Cabbage & 4/5 & & 5/5 & \\
& Apple & 5/5 & & 5/5 & \\
\midrule

\multirow{8}{*}{\specialcell{Daily \\ Items}} & Mustard & 5/5 & \multirow{8}{*}{0.825} & 5/5 & \multirow{8}{*}{0.95} \\
& Spatula & 5/5 & & 5/5 & \\
& Toy Drill & 4/5 & & 5/5 & \\
& Clamp & 2/5 & & 5/5 & \\
& Screw Driver & 3/5 & & 4/5 & \\
& Bottle & 4/5 & & 5/5 & \\
& Wires (Circle) & 5/5 & & 4/5 & \\
& Ropes & 5/5 & & 5/5 & \\

\midrule
\multirow{4}{*}{\specialcell{Translucent \\ Items}} & Cup & 3/5 & \multirow{4}{*}{0.7} & 4/5 & \multirow{4}{*}{0.9} \\

& Sauce Bottle & 2/5 & & 5/5 & \\
& Wine Glasses & 4/5 & & 5/5 & \\
& Green Glasses & 5/5 & & 4/5 & \\

\midrule
\multirow{4}{*}{\specialcell{Complex \\ Items}} & Headphone & 5/5 & \multirow{4}{*}{0.7} & 5/5 & \multirow{4}{*}{0.95} \\

& Sun Glasses & 3/5 & & 5/5 & \\
& Toy Lion & 4/5 & & 4/5 & \\
& Spray Bottle & 2/5 & & 5/5 & \\

\midrule
\multicolumn{2}{c}{Average Success Rate} & & 0.77 & & 0.94\\

\bottomrule
\end{tabular}
\label{table:grasp_results}
\end{table}

\begin{figure}[t]
\centering
\includegraphics[width=\linewidth]{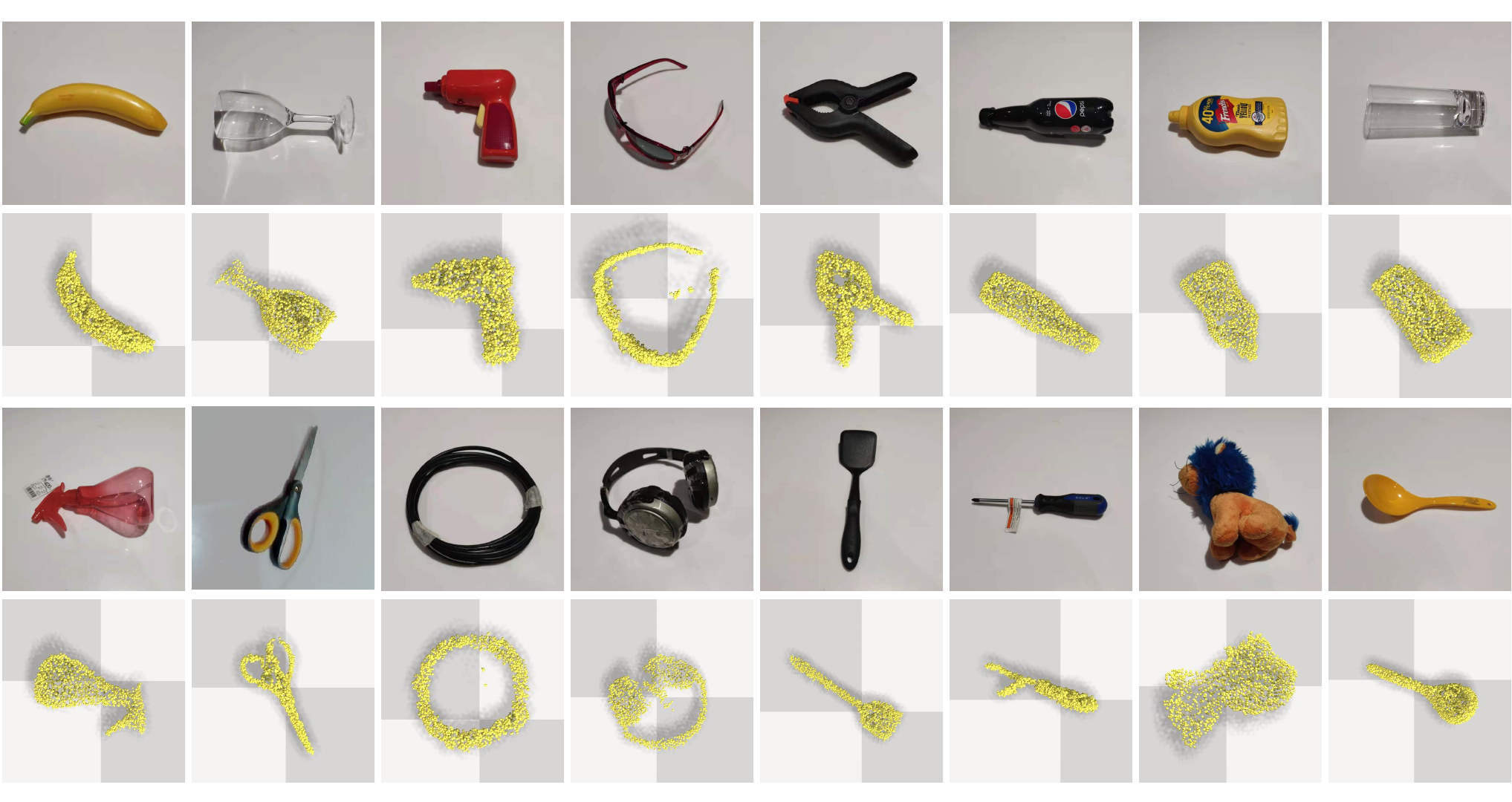}
\caption[HMM]{Real world particle reconstruction and grasping results. Our approach \methodName grasps various novel objects with pure RGB images as input without any offline training. With particle-based object representation, our method connects the image observation with a particle dynamics model and enables model-based planning on novel objects.}
\label{fig:real_reconstruction}
\end{figure}

We evaluate our 3D reconstruction quantitative results on the common benchmark dataset ShapeNet \cite{chang2015shapenet} with the results shown in table \ref{table:reconstruction_results}. Total 8 categories, 50 objects per category, are evaluated. Shown in \tabref{table:reconstruction_results}, our reconstruction performance is not perfect but surprisingly decent. For some categories such as plane, lamp, firearm, and watercraft, our approach achieves better results with the intersection of union (IoU) metrics. It shows that our approach works on complicated objects. However, an interesting observation is that our approach's performance drops significantly on simple object categories such as speaker and couch, often a large box. It is because IoU measures the entire object, including the interior space, which is not meaningful as we do not know whether an object is empty or not given image observations. Our approach generates a point cloud from masking images and has no explicit objective for the interior space. However, other baselines such as Point Set \cite{fan2017point} trains their generator with ground truth point cloud and
hence knows the prior information. During the testing phase,
categorical information is also provided to Point Set, and 3DR2N2 \cite{choy20163d}. We do not claim to outperform the two baselines but to show we can reconstruct the novel objects reasonably well for next-stage object manipulation. We cite them as a performance indicator.

In addition to the deep learning-based approach, we favorably considered other options in the multi-view stereo approach. Typical multi-view stereo methods usually rely on object feature detection or the Lambertian assumption of photo-consistency \cite{furukawa2009accurate, furukawa2010towards}. However, in the real world, transparent objects do not follow the photo-consistency assumption and many other objects have featureless surfaces, leading to a large number of failure cases in traditional multi-view stereo methods. Due to several problems mentioned above, we stick to our own reconstruction method.

Comparing FC-100 and FC-1, FC-100 improves the performance on most categories. CNN-100 achieves a slightly better performance than FC-100 with image inputs as the images provide some prior information that helps with the reconstruction. With the experiment results, we conclude that our 3D reconstruction approach works on novel objects given different views of masking images.


\subsection{Object Grasping}

This experiment aims to answer whether our approach \methodName works on the real robot grasping tasks. We have an eye-in-hand RGB camera that follows a precomputed trajectory to take a short video. To obtain the object's masking image for arbitrary objects, we leverage the pre-trained foreground segmentation algorithm U2Net \cite{Qin_2020_PR}. 

\subsubsection{Experiment Setups}
We deploy our method \methodName on Fetch robot, see figure \ref{fig:robot} top left. The robot has parallel grippers and can perform top-down grasping. We select some objects from the YCB object set \cite{7254318}, and other four translucent objects as well as daily objects. There are 20 objects in total. Figure \ref{fig:robot} left bottom and right show the test set items, and each item is grasped in isolation.

\subsubsection{Evaluation Protocol}
We use the grasping success rate to evaluate the general performance of our approach. There are 20 unseen objects, and we test 5 trials each object with random pose and position. During all the trials, the object remains constant. Each grasp attempt (episode) is a one-time grasping action. If no object has been grasped at the end of the grasping action, the attempt is regarded as a failure.

\subsubsection{Implementation Details}
The robot takes a video of objects and generate masking images using U2Net \cite{Qin_2020_PR} for particle state reconstruction with number of particles $N=2000$ and number of views $I \approx 70$. The predicted particle state is then fed into the SOTA particle based simulator Nvidia-Flex \cite{gameworks2018nvidia} for action planning. We develop the simulator based on PyFlex \cite{PyFleX2019}. Other particle-based simulators such as Taichi \cite{hu2019taichi} and Differentiable Taichi \cite{hu2019difftaichi} are also available. Three 2080Ti GPUs are used for action planning described in section 3C and one 2080Ti for reconstruction.

\begin{table}[t]
\centering
\caption{\small Robot Pushing Results}
\begin{tabular}{ccccc}
\toprule

& \multicolumn{2}{c}{PushNet \cite{Li2018PushNet}} & \multicolumn{2}{c}{\methodName (Ours)} \\
\cmidrule(lr){2-3}\cmidrule(lr){4-5}
                 & Count    & Avg Steps & Count  & Avg Steps   \\
\midrule

Mustard & 10/10 & 5.90 & 9/10 & 4.44 \\
Toy Drill & 9/10 & 6.22 & 10/10 & 3.30\\
Sun Glasses & - & - & 10/10 & 4\\
Banana & 10/10 & 9.80 & 10/10 & 4.30\\
Black Box & 10/10 & 6.00 & 10/10 & 4.40\\
Clamp & 9/10 & 7.11 & 9/10 & 5.11\\
Soup Ladle & 9/10 & 7.56 & 9/10 & 6.11\\
\\[5pt]
\midrule
Average & 57/60 & 7.1 & 67/70 & 4.49\\

\bottomrule
\end{tabular}
\label{table:push_results}
\end{table}

\begin{figure}[t]
\centering
\includegraphics[width=1.0\linewidth]{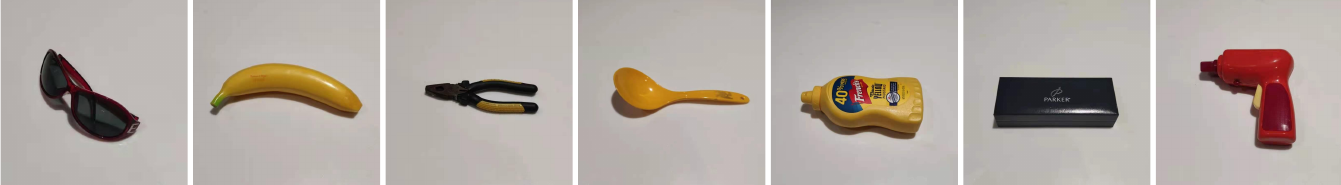}
\caption[HMM]{Objects for pushing tasks. From left to right are: sun glasses, banana, clamp, soup ladle, mustard, black box, toy drill. }
\label{fig:pushing_objects}
\end{figure}

\subsubsection{View Points Selection}

We choose a set of views to maximize the coverage of the target object. In practice, we choose a fixed trajectory to take a short video and perform reconstruction on the collected images. We discover that the top views reveal the most information about the object geometry and the side views, which are very close to the table,  reveal the height information. It indeed matches the three-view drawings convention: front, left and right views. We visualized the view points in \figref{fig:robot}.

The choice of views can be extended to active information gathering that contributes to 3D reconstruction, which we leave for future study. For example, we can formulate the problem of determining the set of views as a problem of “the Next Best View” (NBV) problem and leverage on previously developed NBV solutions \cite{pito1999solution}.

\subsubsection{Results}

We visualize some of the real-world object particle reconstructions in figure \ref{fig:real_reconstruction}. Those items have very different shapes and challenging appearances. For example, the spray bottle and the headphone have irregular shapes, and the transparent wine glasses and cups have a difficult appearance. Suggested by the results, our multi-view particle reconstruction approach can reconstruct the 3D structure reasonably well in the real robot task with a single RGB camera, even with noisy masking images. In real robot experiments, we pipeline the reconstruction process so that each new image collected is immediately available for reconstruction without waiting for the robot to stop. Reconstructing 2000 particles takes 6 seconds to converge, while 1500 particles takes 4 seconds, and the grasp planning takes 8 seconds to generate the required actions. The planning time can be further reduced if we train a policy and use the trained policy to guide the search. We leave this problem for future study. 

\begin{figure*}[t]
\centering
\includegraphics[width=\linewidth]{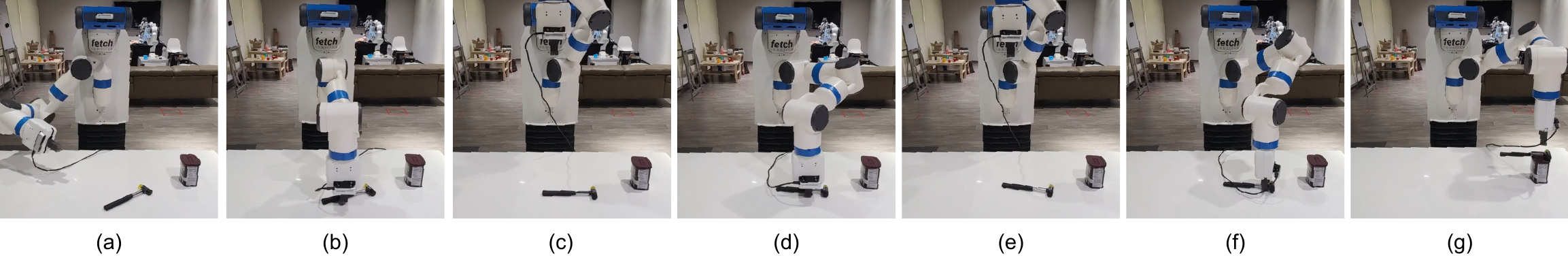}
\vspace{-15pt}
\caption[HMM]{Pushing, grasping, and placing a hammer with an uneven mass distribution. We show the intermediate steps: (a) object reconstruction (b) push (c) position re-estimation (d) push (e) position re-estimation (f) grasp (g) object stacking. The robot manipulates the object interactively to estimate the mass distribution in order to stack the hammer. }
\label{fig:placing}
\end{figure*}

The results are shown in table \ref{table:grasp_results}. The real robot experiment results suggest a success rate of 0.94 over 100 grasping trials for our approach. We compare with DexNet 2.0  \cite{mahler2017dex} that grasps object in isolation, as DexNet 4.0 is for grasping and suction in clutter and DexNet 3.0 is for Suction. The DexNet 2.0 achieves 0.77 success rate in the our experiment. The results suggest our particle model-based planning approach \aname outperforms DexNet 2.0 in all object categories, especially for translucent items and complex items.

Our object segmentation process relies on the foreground segmentation network U2Net \cite{Qin_2020_PR}, and some failures come from noisy object-masking images produced by U2Net, such as failures on translucent green glasses and cups. U2Net is trained on a large set of images. However, it is not explicitly optimized for translucent objects and shadows. Our approach assumes that we can detect object masking images, so in principle, this problem is outside the scope of this work. In fact, we can address the issue in the future by fine-tuning U2Net with translucent objects and shadows. We conclude that our approach \methodName works on the real robot grasping task with complicated object shapes based on the real-world experiment results.

\subsection{Object Pushing}
We further evaluate our approach on the object pushing tasks to find out whether our method generalizes to other object manipulation tasks. Similar to the Figure \ref{fig:robot}, in the pushing experiment we have the same robot setup. Figure \ref{fig:pushing_objects} shows the test objects in the experiment. After each push, the robot takes an additional image to re-estimate the object's current position and configuration with the algorithm described in section 3D.

\subsubsection{Evaluation Protocol}
We use the push success rate and number of steps to evaluate the average performance. We test on 7 novel objects, 10 trials each, with random goal pose and position. Goal positions in x direction are sampled uniformly from (0.1m, 0.2m) $\cup$ (-0.1m, -0.2m) and y direction are sampled uniformly from (-0.15m, -0.25m). Rotational goals are sampled uniformly from (90, 30) $\cup$ (-30, -90). The push task is considered as a success when the object is 5cm closer to the goal with orientation error less than 15 degrees. The maximum allowed number of steps is 25.


\subsubsection{Results}
Table \ref{table:push_results} shows the real robot pushing experiment results. We compare with the modern object pushing baseline PushNet \cite{Li2018PushNet} that trains its policy in a simulator and applies to a real robot. Our approach and PushNet achieve a high success rate, but our approach produces fewer steps to achieve the goal. We leverage on a known particle-based dynamics model and explicit model-based planning, which produces near-optimal actions for each push. Despite the high success rate of PushNet, we observed PushNet often generates sub-optimal actions that incur high costs on difficult objects. Taking the banana as an example, pushing bananas seems like an easy task to complete, but it turns out contrary to expectations. The toy banana is light, which means it rotates easily if the push direction does not accurately go through the center of mass. Sub-optimal push actions hence likely result in large unexpected rotations on the banana. Our approach relies on planning which is good at producing accurate actions and achieves better results on difficult objects \eg banana.

A typical failure case occurs in both methods when the object is pushed out of the robot working area, resulting in unreachable positions. The following actions hence can not be executed, and the task fails. On top of that, the item sunglasses completely fails on PushNet as PushNet uses depth image-based object segmentation that fails on the partially translucent object. Since object segmentation is not the main contribution of PushNet, we exclude the sunglasses result for PushNet.

In summary, our approach naturally generalizes object-pushing tasks by changing the actions space and reward function for the MPC planner.

\subsection{Object Pushing, Grasping, and Placing}
We demonstrate the ability to perform pushing, grasping and placing task all together with our proposed approach. The robot setup is shown in figure \ref{fig:placing}. The system manipulates a hammer with unknown mass distribution, grasps it, and stacks it on a small stand. Without understanding the center of the mass, simply stacking the object's center on the small stand likely leads to a failure. The robot first reconstruct a particle state and then pushes the object for several times to estimate the novel object's center of mass (CoM). The video link can be found in the abstract.

\section{Conclusion}
We present \methodName, a  particle-based object manipulation framework, which enables robots to achieve dynamic manipulation on a variety of tasks, including grasping, pushing, and placing with novel objects. The particular particle state representation bridges the learning for complex observation and planning for decision making, connecting the complex visual observation to a known dynamics model. This idea enables the entire system to deal with complex objects with arbitrary shapes and at the same time has the understanding of the dynamics system for better action planning. We evaluate our approach to real robots with grasping, pushing, and placing tasks, showing the system's effectiveness and generalization to various robot tasks. 

\section*{Acknowledgements}
This research is supported by the National Research Foundation, Singapore under its AI Singapore Programme (AISG Award No: AISG2-PhD-2021-08-015T) and  A\(^*\)STAR under the National Robotics Program (Grant No. 192 25 00054).




\bibliographystyle{plainnat}
\bibliography{references}

\end{document}